\documentclass[a4paper,fleqn]{cas-dc}

\usepackage[numbers]{natbib}

\def\tsc#1{\csdef{#1}{\textsc{\lowercase{#1}}\xspace}}
\tsc{WGM}
\tsc{QE}
\tsc{EP}
\tsc{PMS}
\tsc{BEC}
\tsc{DE}

\usepackage{longtable}
\usepackage{float}

\begin{document}
\let\WriteBookmarks\relax
\def\floatpagepagefraction{1}
\def\textpagefraction{.001}

\shorttitle{}

\shortauthors{ }

\title [mode = title]{Interactive 3D visualization of surface roughness predictions in additive manufacturing: A data-driven framework}

\author[1]{Engin Deniz Erkan}[orcid=0009-0001-1641-1942]

\cormark[1]

\ead{engin.erkan@metu.edu.tr}
\affiliation[1]{organization={Department of Data Informatics, Graduate School of Informatics, Middle East Technical University}, 
   city={Ankara},
     citysep={},
    postcode={06800}, 
    country={Turkey}}

\author[2]{Elif Surer}[orcid=0000-0002-0738-6669]

\ead{elifs@metu.edu.tr}

\affiliation[2]{organization={Department of Modeling and Simulation, Graduate School of Informatics, Middle East Technical University},
   city={Ankara},
     citysep={},
    postcode={06800}, 
    country={Turkey}}

\author[3,4]{Ulas Yaman}[orcid=0000-0001-8598-9281]

\ead{uyaman@metu.edu.tr}

\affiliation[3]{organization={Department of Mechanical Engineering, Middle East Technical University}, 
   city={Ankara},
     citysep={}, 
    postcode={06800}, 
    country={Turkey}}

\affiliation[4]{organization={Welding Technology and NDT Center, Middle East Technical University},
   city={Ankara},
     citysep={},
    postcode={06800}, 
    country={Turkey}}

\cortext[cor1]{Corresponding author} 

\begin{abstract}
Surface roughness in Material Extrusion Additive Manufacturing varies across a part and is difficult to anticipate during process planning because it depends on both printing parameters and local surface inclination, which governs the staircase effect. A data-driven framework is presented to predict the arithmetic mean roughness ($Ra$) prior to fabrication using process parameters and surface angle. A structured experimental dataset was created using a three-level Box--Behnken design: 87 specimens were printed, each with multiple planar faces spanning different inclination angles, yielding 1566 $Ra$ measurements acquired with a contact profilometer. A multilayer perceptron regressor was trained to capture nonlinear relationships between manufacturing conditions, inclination, and $Ra$. To mitigate limited experimental data, a conditional generative adversarial network was used to generate additional condition-specific tabular samples, thereby improving predictive performance. Model performance was assessed on a hold-out test set. A web-based decision-support interface was also developed to enable interactive process planning by loading a 3D model, specifying printing parameters, and adjusting the part's orientation. The system computes face-wise inclination from the model geometry and visualizes predicted $Ra$ as an interactive colormap over the surface, enabling rapid identification of regions prone to high roughness and immediate comparison of parameter and orientation choices.

\end{abstract}


\maketitle

\section{Introduction}
\label{sec1}

Additive manufacturing (AM), commonly referred to as three-dimensional (3D) printing, enables the fabrication of complex geometries through layer-wise material consolidation and has evolved from rapid prototyping toward broader functional and industrial deployment. Nevertheless, surface quality remains a persistent barrier to wider adoption, because layer-based deposition inherently produces surface artifacts such as stair-stepping, waviness, and visible layer lines. These artifacts directly influence functional performance, dimensional fidelity, aesthetics, and the extent of post-processing required to meet application-specific tolerances \cite{SheoranKumar2020, Batuetal2023, RaynaStriukova2016}. In Material Extrusion AM (MEAM), surface roughness is particularly sensitive to geometric features and orientation, which makes it a critical quality attribute for process planning and design for AM workflows \cite{Ahnetal2009, Arnoldetal2019, Galettoetal2021}.

In Fused Filament Fabrication (FFF), a type of MEAM, surface roughness results from complex, nonlinear interactions among controllable process parameters (such as layer height, extrusion temperature, and deposition speeds), material behavior during deposition and cooling, and geometric factors including surface inclination and build orientation. The effect of inclination angle on roughness, primarily due to the stair-stepping mechanism, has been explicitly modeled by representing surface quality as a function of surface angle \cite{Ahnetal2009}. Empirical studies confirm that multiple printer and slicing settings collectively influence surface quality, demonstrating that roughness is rarely determined by a single dominant factor but instead reflects the combined effects of various parameter groups \cite{Arnoldetal2019}. Since process planning often requires balancing quality and efficiency, roughness is typically treated as one objective within broader quality-productivity trade-offs \cite{Galettoetal2021}. Furthermore, post-processing can significantly modify as-printed surfaces; thus, roughness control depends not only on in-process parameter selection but also on available finishing strategies in desktop FFF environments \cite{AlsoufiElsayed2017}.

Despite the practical importance of surface roughness, parameter selection in standard slicing workflows remains largely dependent on experience-based adjustments. Numerous parameters must be defined prior to printing, and the relationships among these settings, orientation, and resulting surface roughness are challenging to predict due to interaction effects and process variability. Most slicer and printer interfaces offer limited predictive feedback regarding surface quality before manufacturing, which promotes trial-and-error experimentation. This approach increases material usage, prolongs iteration cycles, and may result in inconsistent surface quality, especially when technical expertise is limited or when parts feature multiple surfaces with varying inclinations.

A significant practical challenge is that real components seldom exhibit a single representative surface. Instead, they comprise multiple faces with varying local inclinations and orientations, leading to diverse roughness-forming mechanisms and trade-offs within the same part. Therefore, roughness prediction is most effective when interpreted in a geometry-aware context and applied to compare alternative parameter and orientation selections at the part level, rather than as a single scalar outcome.

While previous studies have introduced statistical, learning-based, and optimization-driven methods for roughness modeling, two significant gaps persist in pre-print decision-making for FFF. First, experimental datasets are typically insufficient to span the joint space defined by process parameters and inclination-related geometric descriptors, limiting the generalizability of data-driven predictors. Second, model outputs are often presented in formats that are not readily translatable into practitioner-oriented, geometry-aware tools capable of delivering interpretable, pre-print feedback across complete 3D models.

To address these gaps, this study presents an end-to-end, data-driven methodology for surface roughness prediction and process planning in FFF. The methodology begins with the construction of a systematic experimental dataset using a design-of-experiments approach, which ensures broad coverage of the parameter space and enables robust learning of interaction effects relevant to surface roughness \cite{Galettoetal2021, Bigleteetal2020, Kandananond2021, Pereiraetal2023}. Utilizing this dataset, a multilayer perceptron (MLP) predictor is developed, and conditional synthetic augmentation is investigated through generative modeling techniques informed by research on manufacturing data augmentation and conditional generative adversarial networks (CGANs) \cite{Ekwaro-Osire2025, DouzasBacao2018, Xuetal2019}. In addition to predictive modeling, a web-based decision-support tool is introduced to facilitate practical adoption. The graphical user interface allows users to upload 3D objects and provides real-time visualization of predicted surface roughness on the geometry. By enabling interactive control of printing parameters and build orientation, the interface supports rapid exploration of alternative process plans and immediate assessment of their expected surface-quality outcomes. This interactive workflow facilitates quality-aware parameter selection, enhances interpretability by linking predictions to 3D geometry, and reduces costly trial-and-error cycles by providing pre-print feedback without repeated physical experimentation. Consistent with previous AM decision-support initiatives \cite{ParkTran2017, Aljabalietal2024} and the growing demand for accessible AM workflows across diverse user groups \cite{RaynaStriukova2016}, the proposed system implements data-driven surface-quality prediction as an interactive and geometry-aware process planning tool.

\section{Prior work}
\label{sec2}

Structured experimentation is commonly employed to investigate and model surface roughness dependencies on process parameters and geometry. Design of experiments (DoE) and response surface methodology (RSM) reduce experimental effort while enabling estimation of main effects, interactions, and curvature. Box--Behnken designs have been utilized for surface roughness modeling in FFF, often in conjunction with analysis of variance (ANOVA) to assess factor significance \cite{Bigleteetal2020, Kandananond2021}. DoE and RSM approaches have also been applied to broader print-quality objectives, such as dimensional accuracy, demonstrating the effectiveness of structured experimentation for AM process improvement \cite{Nazanetal2017}. Recent studies have extended these methods to multi-objective contexts that simultaneously consider quality, time, and resource usage, framing parameter selection as a Pareto optimization problem rather than a single-response minimization \cite{Pereiraetal2023}.

Another research direction integrates predictive models with metaheuristic optimization to explore high-dimensional parameter spaces beyond traditional quadratic response surfaces. Hybrid frameworks that combine artificial neural networks (ANNs) with particle swarm optimization (PSO) have been introduced to minimize roughness by optimizing printing parameters using a learned surrogate \cite{Shirmohammadietal2021}. Comparable approaches have incorporated symbiotic organism search (SOS), again leveraging ANN surrogates to facilitate roughness minimization under practical constraints \cite{Saadetal2022}. Related hybrid techniques have also been applied to improve dimensional accuracy by merging modeling and optimization strategies \cite{Deswaletal2019}. Recent studies continue to integrate RSM and ANN for surface-quality modeling, including multiple surface descriptors, and employ desirability-based optimization once predictive accuracy is achieved \cite{Farozeetal2025}. Additionally, genetic algorithm (GA)-optimized hybrid learning has been reported for surface roughness prediction, highlighting sustained interest in evolutionary search methods combined with learning-based predictors for AM quality tasks \cite{AkgunUlkir2024}. Recent AM studies have also introduced hybrid formulations that explicitly integrate data-driven learning with geometric roughness mechanisms, such as stair-stepping, to enhance interpretability while preserving predictive performance \cite{Kugunavaretal2024}.

In addition to DoE-based methods, machine learning (ML) has become a prominent approach for surface roughness prediction, as it can capture nonlinearities and interactions among continuous and discrete variables. Comparative studies of various ML algorithms for FFF-printed parts have shown that geometric descriptors such as wall angle and process variables like layer height are significant determinants of roughness, and that ensemble and MLP models can deliver strong predictive performance \cite{Cerroetal2021}. Other studies focused on ANNs indicate that compact neural models can achieve useful predictive accuracy in limited experimental scenarios \cite{WafaAbdulshahed2021}. Process-specific ML models for extrusion-based AM have also been developed to predict roughness under realistic parameter variability \cite{Lietal2019}. Furthermore, evidence from large-scale AM prediction tasks demonstrates that ML can inform manufacturing decisions when exhaustive experimentation is impractical \cite{Castroetal2021}. Recent reviews confirm the growing importance of roughness prediction, while also highlighting persistent challenges in data availability, generalization, and reproducible measurement protocols \cite{Batuetal2023}. In addition to FFF, process-specific ML methods have been developed for in-situ roughness estimation in laser powder bed fusion, reflecting broader interest in sensor-informed, data-driven surface quality prediction across various AM process families \cite{Toorandazetal2024}.

A major challenge in data-driven AM quality modeling is the limited size and scope of experimental datasets. Surface metrology is labor-intensive, and the input space expands rapidly when accounting for surface inclination, orientation, and multiple process variables. As a result, numerical data augmentation has gained attention as a means to enrich training datasets without increasing the number of physical experiments. Reviews of augmentation methods for numerical manufacturing data highlight that augmentation effectiveness is context-dependent, and inappropriate synthetic sampling can negatively impact predictive performance \cite{Ekwaro-Osire2025}. CGANs have been proposed as a systematic approach for generating condition-specific synthetic samples; empirical studies indicate that CGAN-based generation can enhance learning outcomes compared to traditional oversampling when evaluated appropriately \cite{DouzasBacao2018}. For tabular data, conditional GAN variants have been developed to better capture mixed discrete and continuous variables and complex distributions, enabling more realistic synthetic data generation \cite{Xuetal2019}. Subsequent research has advanced this field by proposing staged or two-stage generation schemes for tabular forecasting tasks, which aim to better preserve dependencies between generated features and the target variable \cite{Moonetal2020}. Additionally, related engineering applications have assessed GAN-based tabular synthesis as a method to enhance predictive performance when experimental data are limited, such as in machine learning-based multiaxial fatigue life prediction \cite{Heetal2022}. In related AM quality monitoring tasks, augmentation is also used to improve robustness in data-limited inspection scenarios \cite{Rachmawatietal2022}, and quality control strategies for augmentation have been introduced to mitigate the risks of harmful or noisy synthetic samples in small-scale industrial defect detection \cite{Faradyetal2024}.

Conditional generative modeling is particularly suitable for AM quality prediction, as the response variable can be explicitly conditioned on controllable printing parameters and pre-print descriptors, such as surface inclination and build orientation. In this framework, CGANs enable the generation of tabular synthetic samples that maintain the joint distribution among process parameters, inclination-related descriptors, and surface roughness, thus enriching sparsely sampled regions of the experimental design without additional physical trials \cite{Ekwaro-Osire2025, DouzasBacao2018, Xuetal2019}. Given that surface inclination is a primary factor in stair-stepping and roughness formation, conditioning the generator on inclination and orientation descriptors ensures that synthetic samples align with process planning decisions and observed roughness patterns \cite{Ahnetal2009, Cerroetal2021}.

The translation of predictive models into practitioner-oriented tools is increasingly recognized as essential for achieving real-world impact, particularly within Industry 4.0 contexts. Decision-support systems have been developed to facilitate the selection of appropriate additive manufacturing (AM) technologies based on specific requirements and constraints \cite{ParkTran2017}. Data-driven process selection frameworks that utilize optimization heuristics, such as genetic algorithms, have also been introduced to recommend AM process options and reduce dependence on expert knowledge \cite{Aljabalietal2024}. These efforts support the broader adoption of AM across diverse user groups and application areas, where accessible and interactive decision support can accelerate iteration cycles and enhance first-time-right outcomes \cite{RaynaStriukova2016}. In related AM domains, practitioner-oriented interfaces have been created to organize and operationalize AM performance knowledge; for example, curated data reviews and database-style design interfaces have been developed for additively manufactured lattice structures to support engineering decision-making \cite{Hanksetal2020}. Nevertheless, for surface-roughness-focused workflows in FFF, a significant challenge persists in integrating predictive modeling with geometry-aware visualization, enabling users to interpret roughness outcomes across entire parts and efficiently explore orientation and parameter alternatives prior to fabrication.

Collectively, these observations motivate an integrated pipeline that incorporates structured experimentation, data-efficient learning, and geometry-aware visualization for pre-print process planning.

\section{Methodology}
\label{sec3}

This study introduces a data-driven framework for predicting surface roughness and optimizing process planning in FFF. The approach integrates systematic experimental data collection, standardized surface metrology, and rigorous data preparation to support learning-based modeling. Surface roughness is estimated using an MLP regressor, which captures nonlinear relationships among printing conditions, surface inclination, and roughness outcomes. To address constraints related to dataset size and representativeness, CGANs generate condition-consistent tabular samples, thereby enriching the training set. The selected predictor is then deployed within a web-based graphical user interface (GUI) that supports interactive inference on user-uploaded 3D geometries by extracting mesh-based inclination descriptors and visualizing predicted roughness as a color-mapped overlay, facilitating real-time exploration of parameter and orientation choices.

\subsection{Design of experiments}
\label{subsec3.1}

A structured experimental design was implemented to quantify the combined effects of key FFF process parameters and surface inclination on arithmetic mean roughness ($Ra$). DoE and RSM are commonly applied in AM surface-quality studies to minimize experimental effort while enabling estimation of main effects, interaction effects, and curvature \cite{Bigleteetal2020,Kandananond2021,Pereiraetal2023}. Inclination-related descriptors are consistently identified as primary contributors to stair-stepping and roughness variation in FFF, which justifies the explicit inclusion of surface angle in the experimental dataset \cite{Ahnetal2009,Cerroetal2021}.

To enhance geometric diversity while maintaining consistency within the part family, a dedicated test specimen with multiple planar faces at distinct inclination angles was used. This specimen comprises 18 measurement surfaces with inclination angles ranging from $0^\circ$ to $170^\circ$ in $10^\circ$ increments, enabling systematic characterization of roughness across a broad spectrum of orientations relevant to practical printing applications.

A three-level Box--Behnken design (BBD) was selected to efficiently explore the process-parameter space. BBD, an RSM-oriented experimental design, enables estimation of second-order response behavior (quadratic terms) and interaction effects while requiring significantly fewer experimental runs than a full factorial design \cite{Bigleteetal2020,Kandananond2021}. Parameter combinations were executed under controlled conditions to isolate the effects of intentional factor variations.

\subsubsection{Selection of process parameters}
\label{subsubsec3.1.1}

Seven process parameters were selected due to established relevance to surface quality and frequent use in prior FFF roughness studies and process-quality investigations \cite{SheoranKumar2020, Batuetal2023, Arnoldetal2019, Galettoetal2021}. Table~\ref{tbl1} summarizes the factors and discrete levels used in the BBD.

\begin{table}[width=\linewidth, pos=H]
\caption{Process parameters and discrete levels used in the Box--Behnken design.}
\label{tbl1}
\begin{tabular*}{\tblwidth}{@{} LLLL@{} }
\toprule
Parameter & Level 1 & Level 2 & Level 3\\
\midrule
Layer Height (mm) & 0.12 & 0.20 & 0.28 \\
Extrusion Temperature ($^\circ$C) & 190 & 200 & 210 \\
Outer Wall Speed (mm/s) & 150 & 200 & 250 \\
Infill Density (\%) & 5 & 15 & 25 \\
Wall Thickness (mm) & 0.36 & 0.42 & 0.48 \\
Bed Temperature ($^\circ$C) & 55 & 60 & 65 \\
Fan Speed (\%) & 60 & 80 & 100 \\
\bottomrule
\end{tabular*}
\end{table}

\subsubsection{Purpose and effectiveness of the Box--Behnken design}
\label{subsubsec3.1.2}

BBD was adopted to obtain broad coverage of the factor space while maintaining experimental feasibility. For $k$ factors at three levels, a full factorial design requires $N = 3^k$, which becomes impractical for physical printing as $k$ increases. With $k=7$, a full factorial design would require $3^7 = 2187$ experimental conditions.

In contrast, the standard Box--Behnken construction uses combinations of mid-level settings to estimate curvature and interaction effects without requiring all corner points. A commonly used run count for BBD is
\begin{equation}
N = 2k(k-1) + C_0,
\label{eq:bbd_runs}
\end{equation}
where $C_0$ is the number of replicated center points. In the present design, $k=7$ yields $2k(k-1)=84$ edge-point combinations.

Three replicated center points ($C_0=3$) were included for two primary methodological reasons. First, center-point replications provide an estimate of pure experimental error, defined as the variability observed when the same factor settings are repeated under nominally identical conditions. This estimate enables statistical assessment of model adequacy and facilitates the distinction between systematic factor effects and measurement or process noise \cite{Bigleteetal2020,Kandananond2021}. Second, center points enhance the stability of curvature estimation because the BBD targets second-order response behavior. Repeated observations near the design center strengthen the model fit in the mid-range region of the factor space and enable a basic lack-of-fit check relative to replicated noise.

In this study, the replicated center-point runs produced nearly identical roughness responses, indicating minimal process variability under fixed hardware and material conditions. Since these runs share identical input settings, including all replicated center-point rows in supervised learning would introduce duplicated feature vectors and potentially bias model fitting. Consequently, for the surface roughness prediction dataset, a single representative center-point observation was retained to avoid duplicated rows, while the remaining measurements were used to confirm repeatability and to verify that measurement noise was limited.

The complete Box--Behnken run matrix and the process settings assigned to each printed specimen are reported in Table~\ref{tbl4} (Appendix). Each object identifier denotes a unique specimen fabricated under the corresponding parameter combination.

\subsubsection{Specimen design for multi-angle surface characterisation}
\label{subsubsec3.1.3}

A dedicated specimen geometry was developed to isolate the effect of surface inclination on roughness, while maintaining all other geometric attributes constant. The specimen comprises 18 planar regions, each offering a sufficiently long measurement track to ensure consistent profilometer acquisition across various angles. The inclination range ($0^\circ$ to $170^\circ$) achieves near-continuous coverage, in contrast to previous studies that typically assess only a limited set of orientations, often confined to near-horizontal and near-vertical surfaces \cite{Ahnetal2009,Cerroetal2021}.

To control for geometric confounders, each inclined region incorporates a flat measurement area, allowing profilometer measurements to be conducted under consistent contact and traverse conditions. This configuration facilitates systematic analysis of the interaction between process parameters and surface inclination in determining roughness outcomes.

\begin{figure}[pos=H]
    \centering
    \includegraphics[width=\linewidth]{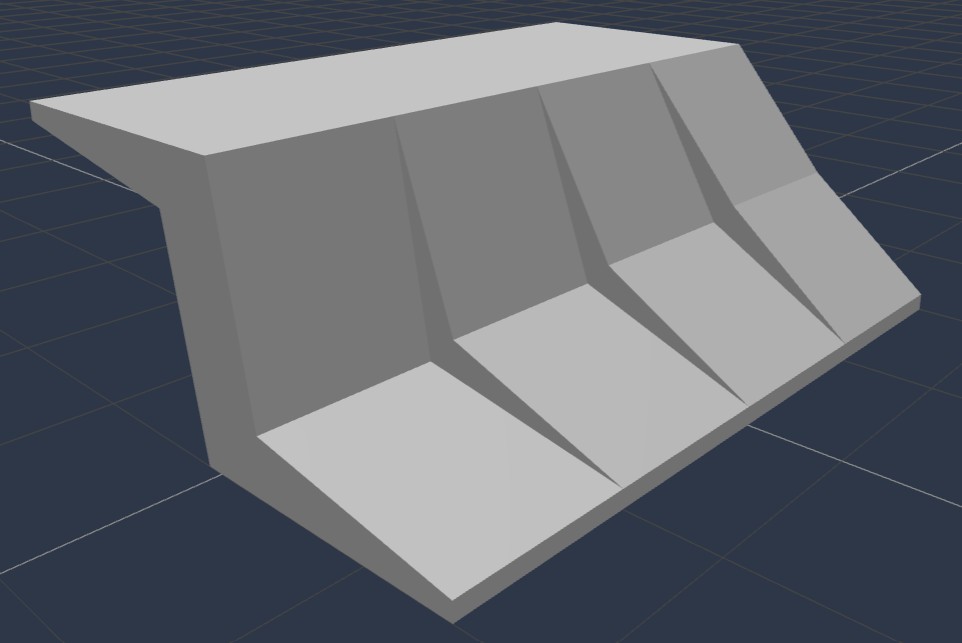}
    \caption{Frontal view of the designed multi-angle specimen.}
    \label{FIG:1}
\end{figure}

\begin{figure}[pos=H]
    \centering
    \includegraphics[width=\linewidth]{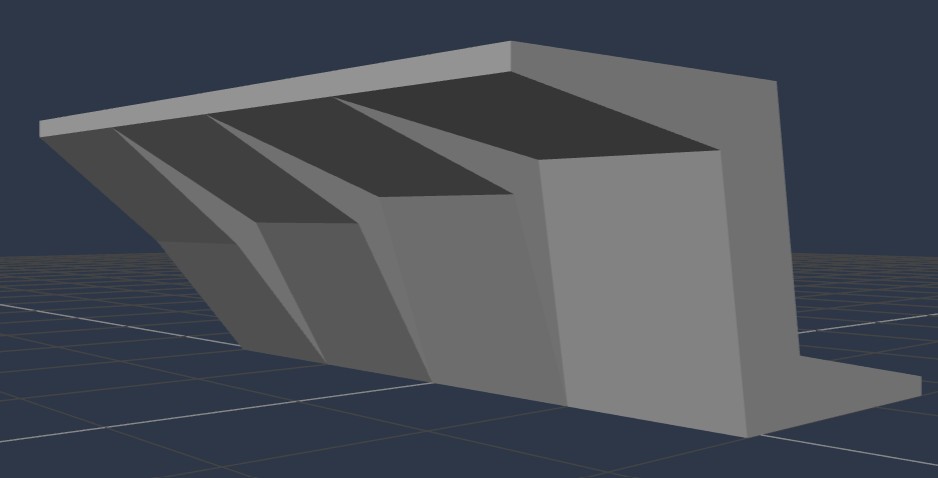}
    \caption{Rear view of the designed multi-angle specimen.}
    \label{FIG:2}
\end{figure}

Figures~\ref{FIG:1} and~\ref{FIG:2} present the frontal and rear views of the specimen. The stepped arrangement ensures clear, repeatable measurement regions while minimizing variations caused by uncontrolled part geometry.

\subsubsection{3D printer and material}
\label{subsubsec3.1.4}

All specimens were fabricated using a Bambu Lab A1 Combo system equipped with a 0.4~mm nozzle. A single filament type was used across all experimental runs to avoid confounding material-to-material variation. The material was polylactic acid (PLA) with a nominal filament diameter of 1.75~mm and a marble color variant. Only the parameters specified by the experimental design were varied; all other machine settings and environmental conditions were held constant across runs to support comparability and isolate factor effects. Crucially, all specimens were printed without support structures to ensure that surface roughness measurements reflected only the interaction between process parameters and surface inclination. This approach prevented surface artifacts or localized damage typically associated with support contact, thereby preserving the integrity of the as-printed surface texture for metrological evaluation.

\subsubsection{Printed objects}
\label{subsubsec3.1.5}

\begin{figure}[pos=H]
    \centering
    \includegraphics[width=\linewidth]{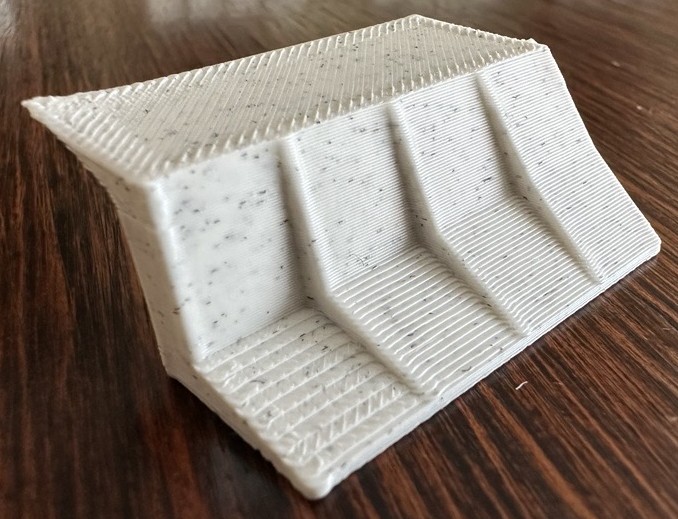}
    \caption{Fabricated specimen.}
    \label{FIG:3}
\end{figure}

\begin{figure}[pos=H]
    \centering
    \includegraphics[width=\linewidth]{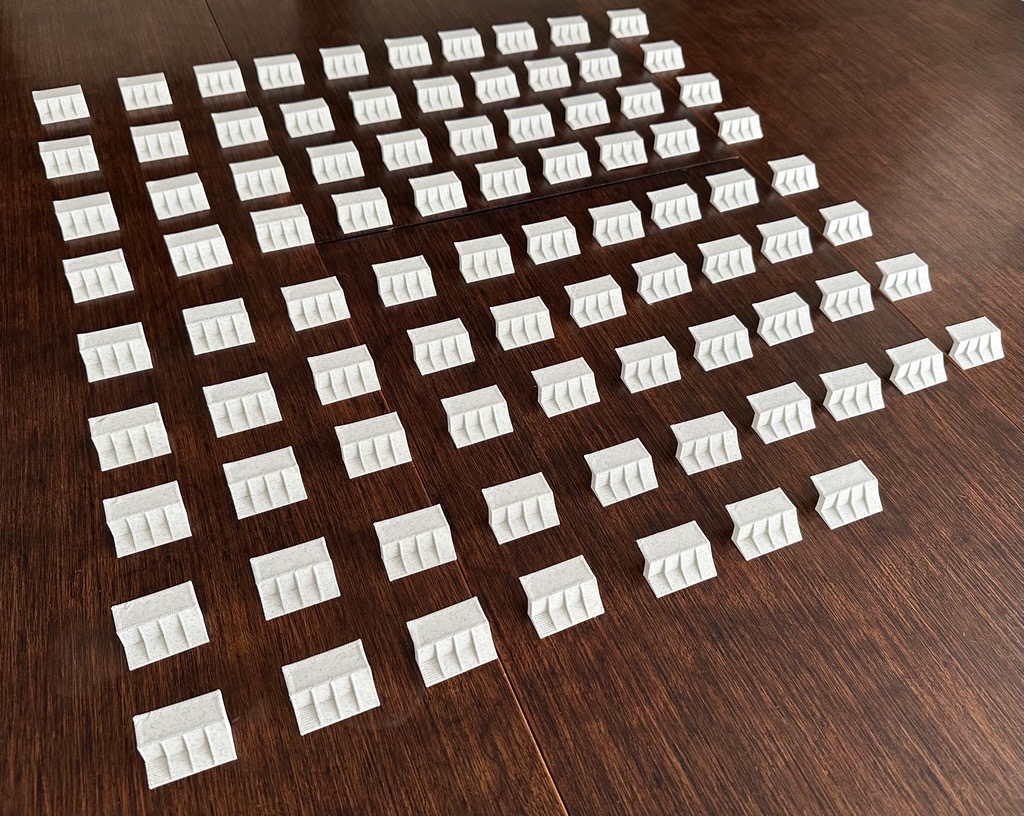}
    \caption{Full set of fabricated specimens corresponding to the Box--Behnken runs.}
    \label{FIG:4}
\end{figure}

A total of 87 specimens were produced corresponding to the 87 Box--Behnken experimental conditions. A representative view of the specimen is shown in Figure~\ref{FIG:3}. The full experimental set is shown in Figure~\ref{FIG:4}.

\subsubsection{Measurement equipment and setup}
\label{subsubsec3.1.6}

Surface roughness was measured using a contact stylus profilometer. This technique enables standardized surface texture evaluation and supports consistent filtering and parameter extraction in accordance with international standards. The measurement configuration was kept constant across all specimens and surfaces to ensure that observed roughness variations resulted from process parameters and surface inclination, rather than from changes in metrology settings. In total, 1566 measurements were taken from different surfaces. To ensure high measurement accuracy and minimize vibration-induced noise, each specimen was securely stabilized on a leveled surface plate during acquisition. This setup maintained stable contact between the stylus and the measurement area, prevented slipping or lateral displacement, especially on steeply inclined surfaces, and supported consistent traverse conditions.

A phase-correct Gaussian filter was applied to separate roughness from the primary profile, in accordance with DIN EN ISO 16610-21 and ASME B46.1. A cut-off wavelength of $\lambda_c = 0.8~\mathrm{mm}$ was used for all measurements. The use of a phase-correct filter prevents phase distortion and minimizes the risk of shifting peaks and valleys, particularly on FFF surfaces that display periodic ridge structures.

\begin{figure}[pos=H]
    \centering
    \includegraphics[width=\linewidth]{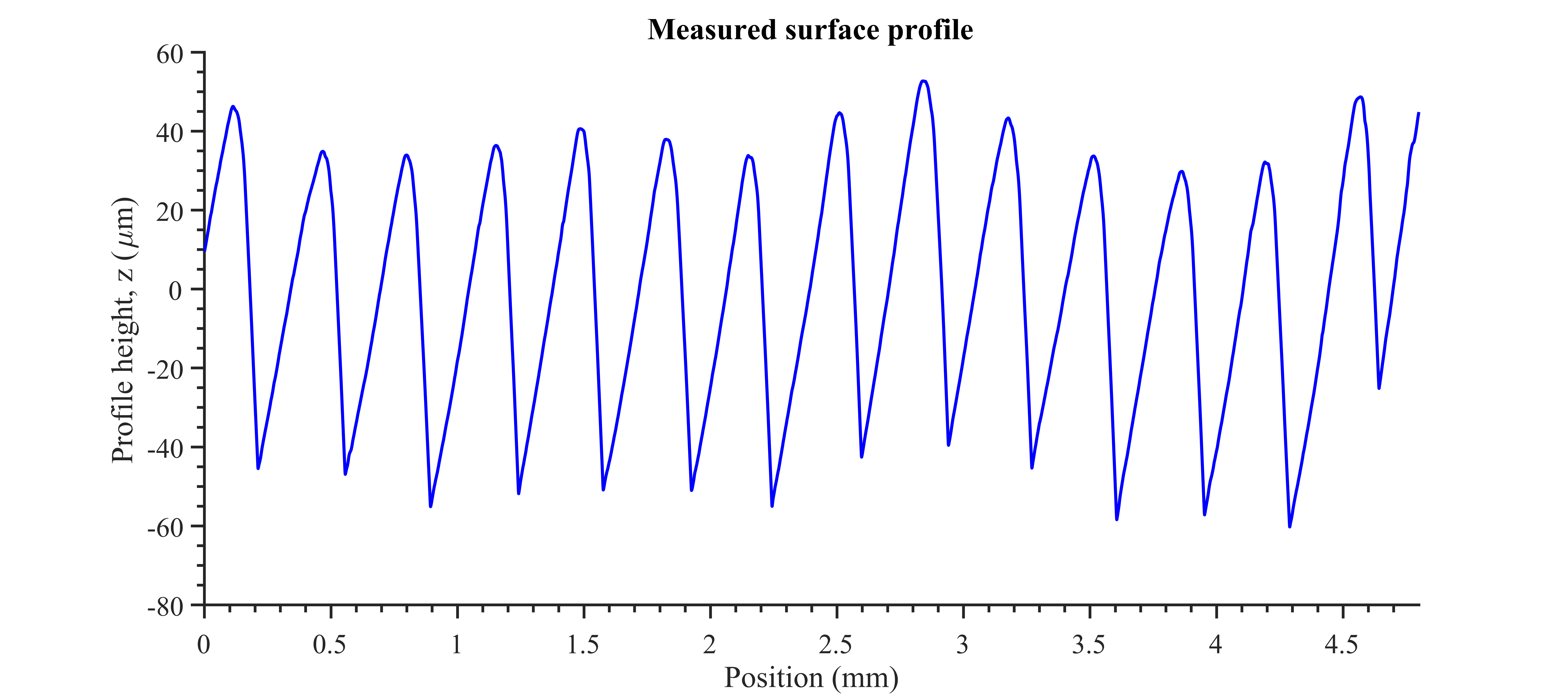}
    \caption{Representative surface profile measured by the stylus profilometer and visualized in MATLAB.}
    \label{FIG:5}
\end{figure}

Measurements were conducted under controlled laboratory conditions to minimize environmental variability. Figure~\ref{FIG:5} presents a representative surface profile acquired by the contact stylus profilometer and replotted in MATLAB. The profile is shown as profile height $z$ ($\mu$m) versus scan position (mm), providing a direct visualization of the measured surface texture along the measurement trace.

\subsection{Surface roughness prediction using an MLP regressor}
\label{subsec3.2}

Surface roughness in FFF results from nonlinear interactions among process parameters, such as layer height, thermal conditions, and deposition speed, as well as geometry-related factors, such as surface inclination. These coupled effects are challenging to capture using low-order parametric models, particularly when multiple variables and interaction pathways are involved. As a result, data-driven regression models have become prevalent for predicting roughness in material-extrusion AM, with artificial neural networks often cited as effective nonlinear surrogates for mapping process inputs to roughness outcomes \cite{Batuetal2023,Kandananond2021,Shirmohammadietal2021,Saadetal2022,Farozeetal2025,Cerroetal2021,WafaAbdulshahed2021}. In alignment with this approach, an MLP regressor was chosen as the predictive model based on its capacity to approximate complex nonlinear functions from tabular data, its compatibility with standardized feature scaling and mixed process descriptors, and its suitability for the relatively small-to-moderate experimental datasets typical in AM surface metrology studies \cite{Batuetal2023,Cerroetal2021}.

The MLP receives as input a tabular feature vector comprising the printing process parameters (Table~\ref{tbl1}) and the surface inclination descriptor, and outputs a scalar estimate of the arithmetic mean roughness $Ra$ (in $\mu$m). The network architecture consists of multiple fully connected layers, each followed by batch normalization and a nonlinear activation function. Dropout is used to address overfitting, a significant concern in experimental AM datasets with limited unique printing conditions. The output layer employs a linear activation function to enable continuous-valued regression. Model training utilizes the Adam optimizer with gradient clipping to stabilize optimization in the presence of noisy gradients and heterogeneous feature scales.

\subsubsection{Data splitting and evaluation protocol}
\label{subsubsec3.2.1}

A rigorous evaluation protocol was implemented to ensure an unbiased assessment of generalization performance and to mitigate optimistic results due to information leakage. Initially, a hold-out test set comprising 15\% of the available samples was set aside and excluded from model selection. This test set remained constant throughout development and was used solely for final performance evaluation. The remaining 85\% of the data constituted the training and validation pool.

Model selection and hyperparameter optimization were conducted using five-fold cross-validation (CV) on the training and validation pool. In each fold, the training subset was used to fit the feature scaler and learn network parameters, while the corresponding validation subset was used for early stopping and calculating validation metrics. To prevent information leakage, min-max normalization was applied within each fold by fitting the scaler to the fold's training subset and subsequently transforming the fold's validation subset with the same scaler. Following hyperparameter selection, the final MLP model was retrained on the entire training and validation pool with the selected configuration and evaluated once on the hold-out test set.

\subsubsection{Hyperparameter optimization}
\label{subsubsec3.2.2}

Hyperparameter optimization was performed using Optuna, which employs efficient sequential model-based search strategies, such as Tree-structured Parzen Estimator (TPE) sampling, for black-box optimization of model configurations. The optimization objective was the mean validation mean absolute error (MAE) across the five CV folds, serving as the primary error metric for model selection. The search space encompassed the number of hidden layers, layer widths, dropout rate, activation function, learning rate, $\ell_2$ regularization strength, and batch size.

For each fold, training incorporated early stopping based on validation MAE to mitigate overfitting and prevent unnecessary epochs after validation error plateaued. A learning rate scheduler with reduce-on-plateau functionality was implemented to improve convergence when progress slowed. To increase computational efficiency, median-based pruning was applied at the fold level, enabling early termination of underperforming trials when intermediate results were unlikely to surpass the current best configurations. Upon completion of optimization, the optimal hyperparameters were selected to train the final MLP regressor.

\subsubsection{Performance metrics}
\label{subsubsec3.2.3}

Model performance was quantified using MAE, mean squared error (MSE), coefficient of determination ($R^2$), and mean absolute percentage error (MAPE). These metrics evaluate the predictive accuracy of the models by comparing the experimentally measured surface roughness values against the values predicted by the learning algorithms across the held-out test set. MAE was selected as the primary metric because it provides a direct and interpretable error magnitude in the same units as the response variable ($\mu$m), which aligns with established surface metrology practices and engineering tolerances. Unlike MSE, MAE does not excessively penalize infrequent large errors due to squaring, making it less sensitive to rare outliers resulting from measurement variability or local surface artifacts. $R^2$ is reported to indicate the proportion of variance explained by the model; however, it is not used as the sole selection criterion because it is sensitive to the variance of the target distribution and does not convey an absolute error scale. MAPE is included to offer a relative error perspective, but its instability when measured roughness values approach zero necessitates its use as a supplementary metric rather than as the optimization objective. Reporting MAE, MSE, $R^2$, and MAPE together is consistent with established practices in learning-based AM surface-quality prediction studies and facilitates robust comparison across modeling approaches \cite{Batuetal2023,Shirmohammadietal2021,Saadetal2022,Farozeetal2025,Cerroetal2021}.

\subsection{Data augmentation via CGAN model}
\label{subsec3.3}

Experimental datasets for AM surface-quality modeling are frequently constrained in both size and coverage due to the time-intensive nature of fabrication and profilometry. The input space expands rapidly with the inclusion of multiple process variables and surface inclination, resulting in sparsely sampled regions and diminished generalization of data-driven predictors. Numerical augmentation has been investigated as a means to enrich training distributions without a proportional increase in physical experimentation. However, the effectiveness of augmentation is highly context-dependent and requires rigorous evaluation to prevent performance degradation \cite{Ekwaro-Osire2025}. Within this context, CGANs offer a systematic approach to generate synthetic samples under specified conditions by learning the conditional distribution of the target variable given the input descriptors \cite{DouzasBacao2018,Xuetal2019}.

\begin{figure}[pos=H]
    \centering
    \includegraphics[width=\linewidth]{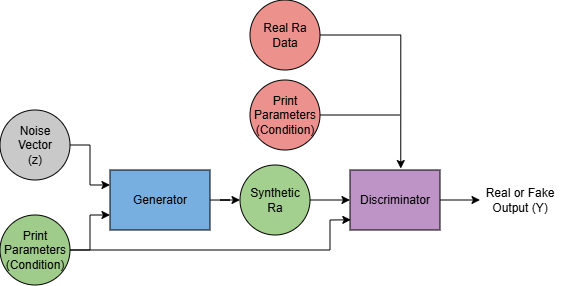}
    \caption{Schematic illustration of the CGAN-based tabular augmentation workflow.}
    \label{FIG:6}
\end{figure}

Within this framework, CGAN-based augmentation is employed to generate tabular synthetic data that aligns with the experimental feature space. The conditioning vector comprises scaled process parameters and the surface inclination descriptor, while the generator produces a synthetic roughness value ($Ra$) conditioned on these inputs. The discriminator receives pairs of $(Ra,\mathbf{x})$ and learns to distinguish real from synthetic conditional samples, which encourages the generator to produce outputs consistent with the specified conditions. This approach is consistent with tabular CGAN methodologies developed for mixed and multimodal numerical data, where conditional generation is intended to preserve feature dependencies rather than generate unconditional samples \cite{Xuetal2019}. A schematic overview of the conditional generation and discrimination workflow used in this study is shown in Figure~\ref{FIG:6}.

To ensure a rigorous and unbiased comparison with the predictor, the hold-out test set remained fixed and was excluded from all stages of CGAN training, augmentation, and model selection. The CGAN was trained solely on the real training subset to prevent information leakage from validation or test data into the synthetic generation process. Following training, synthetic samples were produced by sampling condition vectors from the empirical training distribution and introducing a stochastic perturbation in the normalized feature space. This approach facilitates exploration of local neighborhoods around observed conditions and enhances diversity while maintaining physical plausibility within the scaled feature ranges. The synthetic samples were subsequently combined with the real training data to create augmented training sets at various augmentation ratios, enabling analysis of the impact of training-set enlargement on MLP predictive performance.

The following evaluation investigates whether synthetic enrichment improves the generalization capability of the MLP regressor described in Section~\ref{subsec3.2}. A fixed hold-out test set, comprising 15\% of the full dataset, was reserved and excluded from all CGAN training, augmentation, and model selection procedures. The remaining 85\% of the data was divided into a real training subset (70\%) and a real validation subset (15\%) prior to augmentation. For each augmentation ratio, the MLP was trained on the augmented training subset, while the validation subset remained exclusively real and unchanged. Model selection relied solely on validation MAE to prevent inadvertent optimization toward test performance. The hold-out test set was accessed only once for final reporting of MAE, MSE, $R^2$, and MAPE under the selected configuration. This protocol ensures that observed performance differences can be attributed to augmentation effects rather than changes in evaluation data or selection criteria.

Since synthetic data may introduce bias if considered as reliable as physical measurements, the training procedure applies reduced sample weights to synthetic observations during MLP fitting. This weighting approach aligns with recommendations that augmentation should not overshadow empirical data, particularly when synthetic samples may contain model-induced artifacts. It achieves a balance between expanding data coverage and maintaining fidelity to measured data \cite{Ekwaro-Osire2025}. Additionally, distributional diagnostics were conducted to compare real and synthetic $Ra$ values within the training set, ensuring that synthetic generation produced plausible roughness ranges. Related research in industrial defect learning also emphasizes that augmentation should be accompanied by quality control measures to mitigate the risk of introducing harmful synthetic samples \cite{Faradyetal2024}.

\subsection{Development of the visual decision-support model}
\label{subsec3.4}

Although scalar roughness indicators such as arithmetic mean roughness ($Ra$) provide objective quantification of surface quality, a single numerical value is insufficient for process planning of complex parts. In FFF, surface quality varies spatially because local inclination changes across the geometry and interact with printing conditions, resulting in region-specific roughness. To address this limitation, an interactive visual decision-support model was developed to convert pointwise predictions into a geometry-aware surface map. This model connects tabular inference with intuitive interpretation, enabling users to identify critical regions, evaluate orientation-driven changes, and iteratively refine process parameters before fabrication.

\begin{figure}[pos=H]
    \centering
    \includegraphics[width=\linewidth]{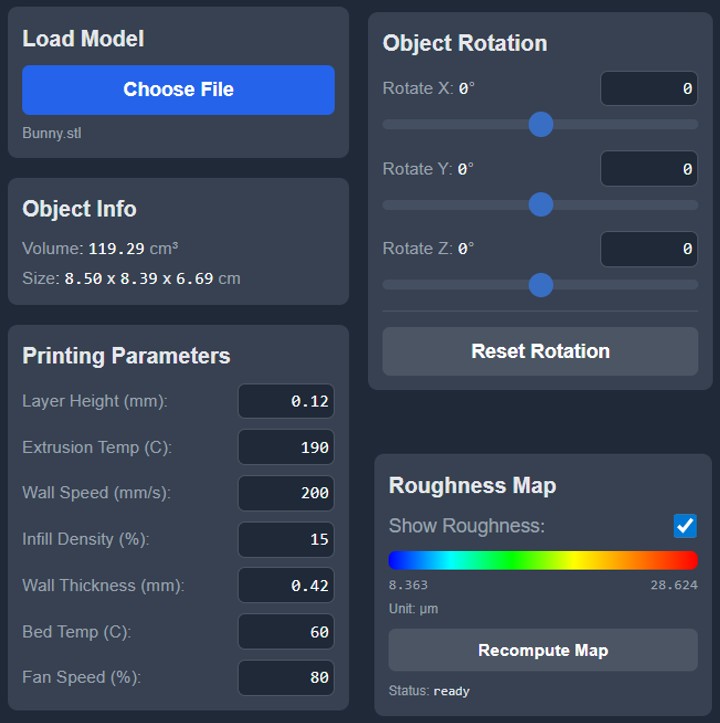}
    \caption{Web-based interface for model upload, parameter input, and three-axis orientation control}
    \label{FIG:7}
\end{figure}

\begin{figure}[pos=H]
    \centering
    \includegraphics[width=\linewidth]{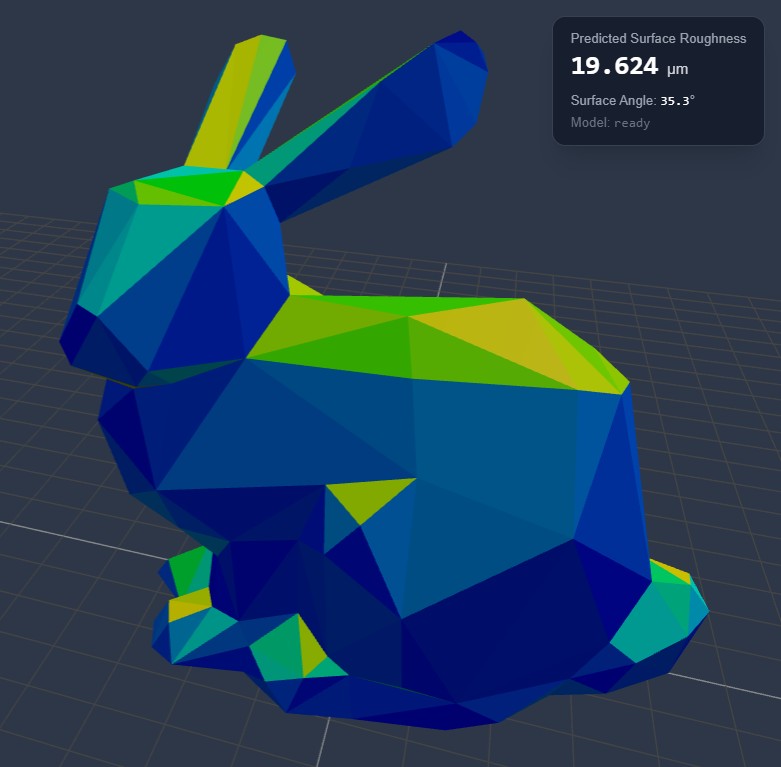}
    \caption{Predicted $Ra$ colormap rendered on the 3D model, with facet-level values via click selection.}
    \label{FIG:8}
\end{figure}

After model selection, the best-performing predictor is implemented as the inference engine within a web-based interface (Fig.~\ref{FIG:7}). The workflow begins with the upload of a user-supplied model (e.g., STL, OBJ, or 3DS), which is rendered interactively in the browser. The interface parses the mesh and calculates local surface inclination using facet normals relative to the build direction. For each surface element, the inclination angle is determined and combined with user-defined printing parameters to form the model input vector. The predictor estimates $Ra$ for each surface element, generating a dense roughness field across the entire geometry. The interface offers both global and local interpretability: users can select any surface region to access the predicted roughness and its associated inclination angle (Fig.~\ref{FIG:8}), enabling inspection at the level of individual facets rather than relying solely on aggregate statistics.

To support rapid interpretation, predictions are visualized on the mesh as a continuous colormap overlay, transforming numerical outputs into an immediate spatial preview of expected surface quality (Fig.~\ref{FIG:8}). This approach maintains the original mesh geometry and visualizes predicted roughness as a colormap overlay, effectively highlighting gradients and hotspots. Users can rotate, zoom, and pan the model to examine all surfaces, and the consistent color scale allows direct comparison of alternatives within a single session.

If the predicted surface roughness does not meet the desired target, users can adjust printing parameters or modify build orientation by rotating the model about the three principal axes (Fig.~\ref{FIG:7}). Changes in orientation alter the build direction and, consequently, the inclination distribution across the mesh; the interface recalculates inclinations and updates the roughness map accordingly. Because inference is interactive, users can immediately observe how parameter adjustments and reorientation affect roughness distribution, making trade-offs explicit, such as improving one region at the expense of another. This immediate feedback loop enables efficient joint exploration of orientation and process settings without repeated fabrication, reducing reliance on costly trial-and-error and supporting surface-quality-aware process planning.

\section{Experimental results}
\label{sec4}

This section reports the experimental findings obtained from the fabricated dataset and provides an exploratory analysis of the measured surface roughness response. The aim is to characterize the roughness distribution and to reveal dominant empirical trends with respect to key variables such as layer height and surface inclination, thereby motivating the subsequent predictive and interpretability results presented in the following subsections.

\subsection{Exploratory data analysis}
\label{subsec4.1}

An exploratory data analysis (EDA) was conducted to summarize the measured surface roughness values, verify data integrity, and examine the empirical relationships between key parameters and the target variable. The dataset comprises 1566 surface measurements collected from 87 printed specimens. Each observation contains eight numerical inputs (seven process parameters and surface angle) and one response variable, the arithmetic mean roughness $Ra$.

All variables were inspected for completeness and consistency. The input space is discrete by construction, since each process parameter is evaluated at three preset levels and surface angle spans $0^\circ$--$170^\circ$ in $10^\circ$ increments. This discretization yields clustered response behavior and motivates the use of nonlinear predictive models in subsequent analyses.

\subsubsection{Dataset structure and descriptive statistics}
\label{subsubsec4.1.1}

Table~\ref{tbl2} summarizes descriptive statistics of $Ra$. The range and dispersion indicate sufficient variability for supervised regression, while the low skewness suggests limited asymmetry; deviations from normality are instead consistent with the multimodal shape observed in Fig.~\ref{FIG:9}.

\begin{table}[width=\linewidth, pos=h]
\centering
\caption{Descriptive statistics of the target variable $Ra$ (in $\mu$m).}
\label{tbl2}
\begin{tabular*}{\tblwidth}{@{} LLLL@{} }
\toprule
\textbf{Statistic} & \textbf{Value} \\
\midrule
Mean & 20.62 \\
Standard deviation & 8.25 \\
Minimum & 2.69 \\
1st quartile (Q1) & 14.21 \\
Median & 20.11 \\
3rd quartile (Q3) & 26.76 \\
Maximum & 46.34 \\
Skewness & 0.31 \\
\bottomrule
\end{tabular*}
\end{table}

\subsubsection{Target variable distribution}
\label{Section4.1.2}

\begin{figure}[pos=H]
    \centering
    \includegraphics[width=\linewidth]{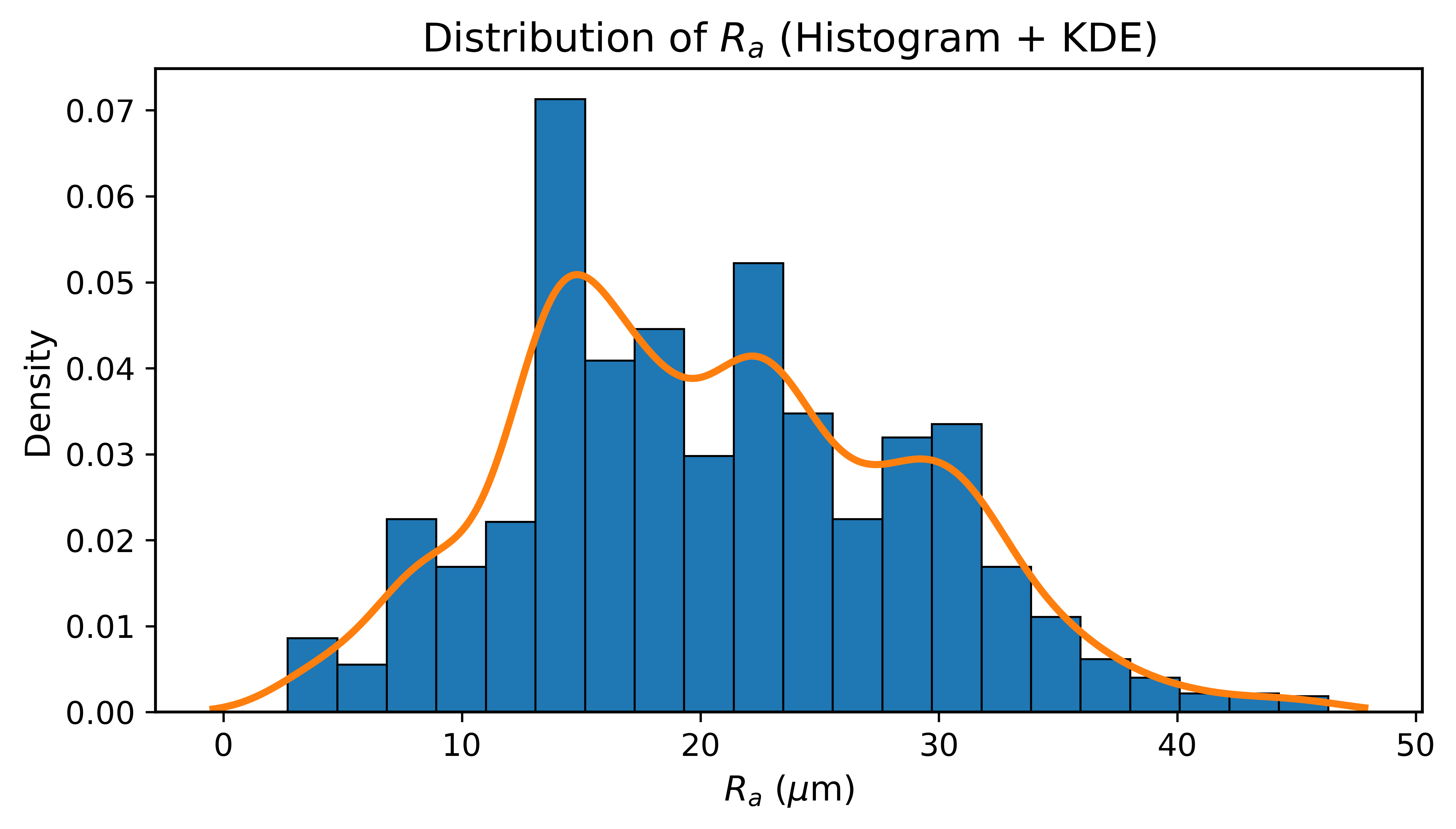}
    \caption{Distribution of surface roughness ($Ra$) shown as a histogram with kernel density estimate.}
    \label{FIG:9}
\end{figure}

The marginal distribution of $Ra$ is illustrated in Fig.~\ref{FIG:9} using a histogram and kernel density estimate. The distribution exhibits multiple modes, consistent with the presence of distinct roughness regimes, likely reflecting combinations of discrete process settings and geometric conditions.

\subsubsection{Feature--target relationships}
\label{Section4.1.3}

\begin{figure}[pos=H]
    \centering
    \includegraphics[width=\linewidth]{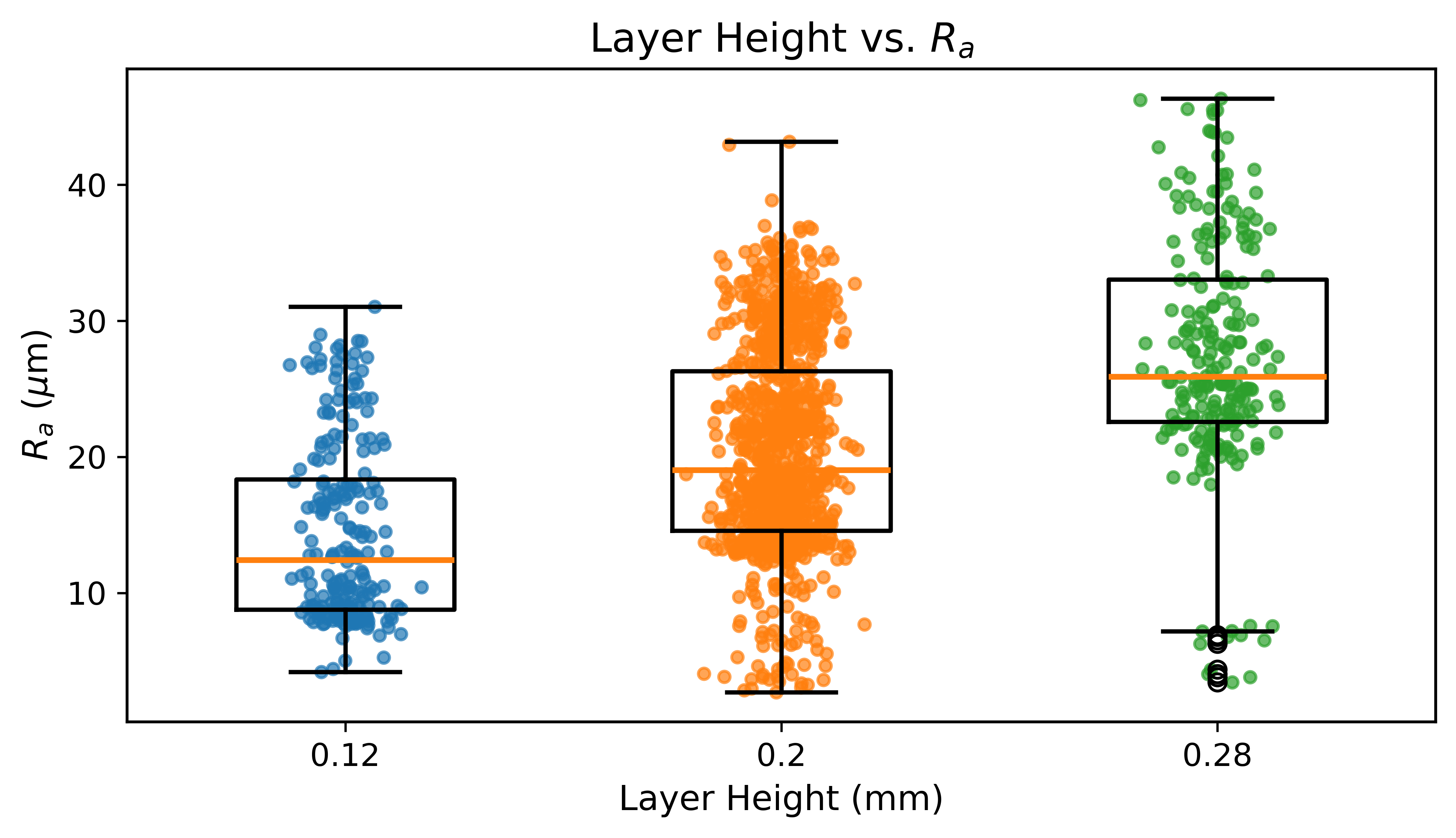}
    \caption{Layer height versus $Ra$.}
    \label{FIG:10}
\end{figure}

\begin{figure}[pos=H]
    \centering
    \includegraphics[width=\linewidth]{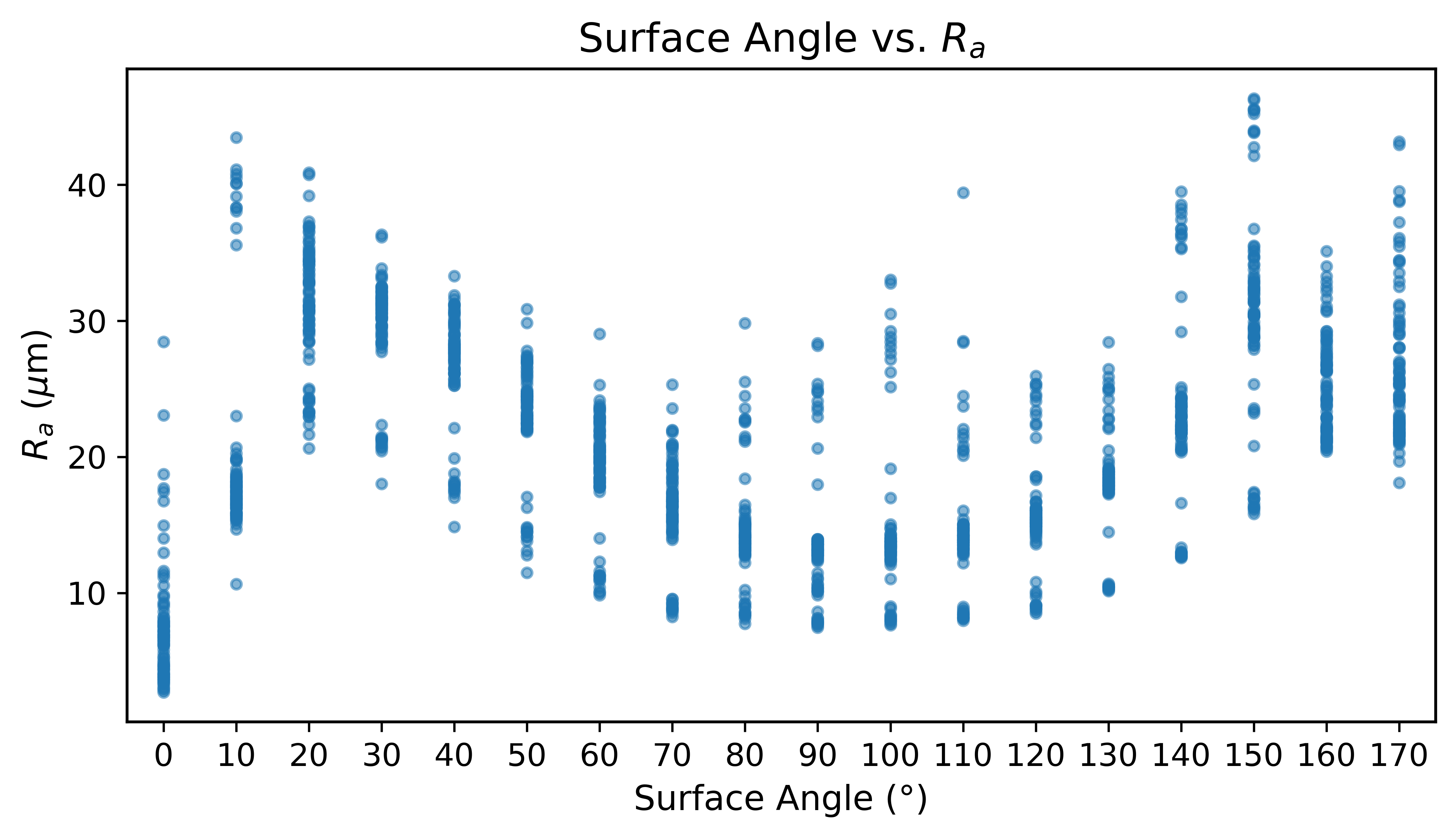}
    \caption{Surface angle versus $Ra$.}
    \label{FIG:11}
\end{figure}

Two dominant empirical relationships are highlighted. First, $Ra$ increases with layer height (Fig.~\ref{FIG:10}), reflecting the increased prominence of layer-wise stair-stepping as the step size grows. Second, $Ra$ shows a pronounced orientation dependence across surface angle levels (Fig.~\ref{FIG:11}), with a clearly non-monotonic pattern, suggesting that angle-sensitive effects should be represented explicitly in predictive modeling.

\subsubsection{Summary of EDA findings}
\label{Section4.1.4}

Overall, the EDA indicates that $Ra$ is governed by both direct process effects (e.g., layer height) and geometry-driven effects (surface angle). The multimodal structure of the target distribution, together with the nonlinearity observed in feature--target relations, supports the use of nonlinear learners and motivates evaluating tabular data augmentation strategies in subsequent sections.

\subsection{MLP results with and without CGAN-based tabular augmentation}
\label{subsec4.2}

This subsection reports the predictive performance of the MLP regressor trained (i) on the experimental dataset only and (ii) on CGAN-augmented tabular training sets. In all cases, the test set is a fixed hold-out set and is used only for final reporting. The augmentation ratio was selected using the validation set by comparing multiple candidate ratios and choosing the configuration that minimized validation error. The best-performing model was obtained when the training set was augmented with a synthetic dataset three times the size of the original training set.

\begin{figure}[pos=H]
    \centering
    \includegraphics[width=\linewidth]{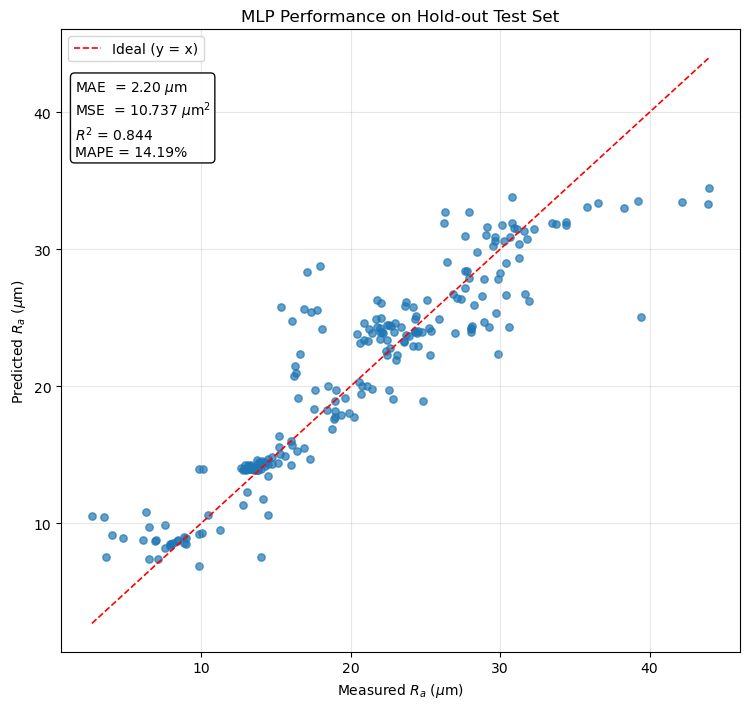}
    \caption{Measured versus predicted $Ra$ on the hold-out test set for the MLP trained with real data only.}
    \label{FIG:12}
\end{figure}

Figure~\ref{FIG:12} presents measured versus predicted $Ra$ values on the hold-out test set for the MLP trained with real data only. The model shows a strong agreement with the ideal $y=x$ trend, indicating that nonlinear relations between process parameters, surface inclination, and roughness can be learned from the available measurements. However, a wider dispersion is observed at higher roughness levels, suggesting increased prediction uncertainty in rougher regimes.

\begin{figure}[pos=H]
    \centering
    \includegraphics[width=\linewidth]{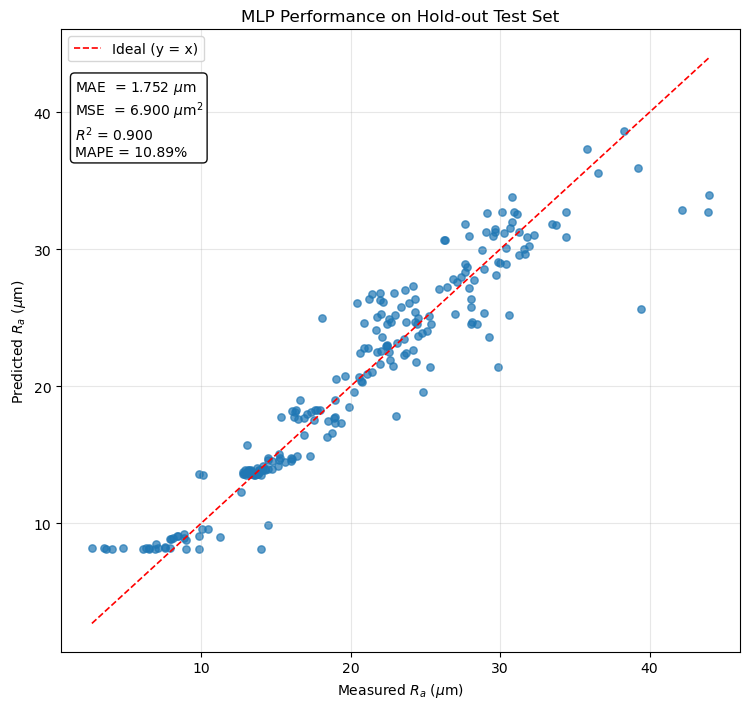}
    \caption{Measured versus predicted $Ra$ on the hold-out test set for the MLP trained with CGAN-augmented tabular data.}
    \label{FIG:13}
\end{figure}

Figure~\ref{FIG:13} shows the corresponding test-set scatter for the selected CGAN-augmented model. Compared to the real-only model, the augmented model exhibits a tighter concentration around the ideal line, reflecting improved generalization on unseen real measurements.

Table~\ref{tbl3} presents the test-set metrics for both experimental settings. Compared to training exclusively on real data, CGAN-based augmentation reduces the MAE and increases the explained variance ($R^2$). These findings suggest that conditional synthetic data enrichment offers advantages in data-limited manufacturing regression, assuming model selection is conducted on a validation set and the test set is strictly reserved for final evaluation.

\begin{table}[width=\linewidth, pos=H]
\caption{Hold-out test performance of the MLP regressor trained with experimental data only and with combined experimental and CGAN-synthesized data.}
\label{tbl3}
\centering
\footnotesize
\begin{tabular*}{\tblwidth}{@{\extracolsep{\fill}} lcccc @{}}
\toprule
\textbf{Training data} & \textbf{MAE} & \textbf{MSE} & \textbf{$R^2$} & \textbf{MAPE} \\
 & ($\mu$m) & ($\mu$m$^2$) &  & (\%) \\
\midrule
Experimental only & 2.200 & 10.737 & 0.844 & 14.19 \\
Experimental + synthetic & 1.752 & 6.900 & 0.900 & 10.89 \\
\bottomrule
\end{tabular*}
\end{table}

Analysis of the validation-based ratio sweep indicates that increasing the augmentation ratio beyond an optimal threshold degrades model performance. This outcome is anticipated, as a CGAN approximates the true conditional data distribution. At moderate augmentation ratios, synthetic samples enhance coverage of sparsely sampled regions and reduce model variance. However, at higher ratios, the training set may become dominated by synthetic patterns, which can amplify minor generator biases or artifacts. Consequently, the regressor may overfit to synthetic-specific structures, resulting in increased error on real test data. Overall, the results demonstrate that CGAN-based augmentation improves MLP performance when used in a controlled manner and when the augmentation ratio is selected using validation data.

\begin{figure}[pos=H]
    \centering
    \includegraphics[width=\linewidth]{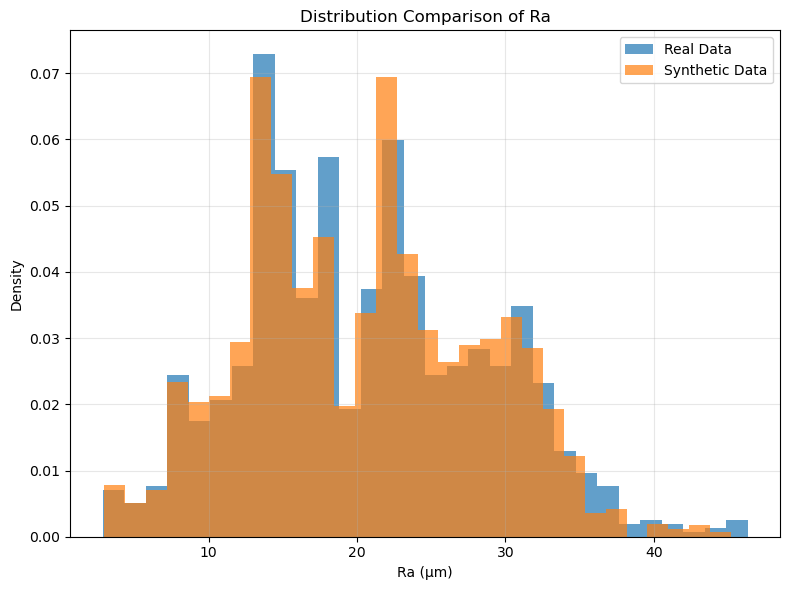}
    \caption{Distribution comparison of $Ra$ values between real training data and CGAN-generated synthetic samples.}
    \label{FIG:14}
\end{figure}

\begin{figure}[pos=H]
    \centering
    \includegraphics[width=\linewidth]{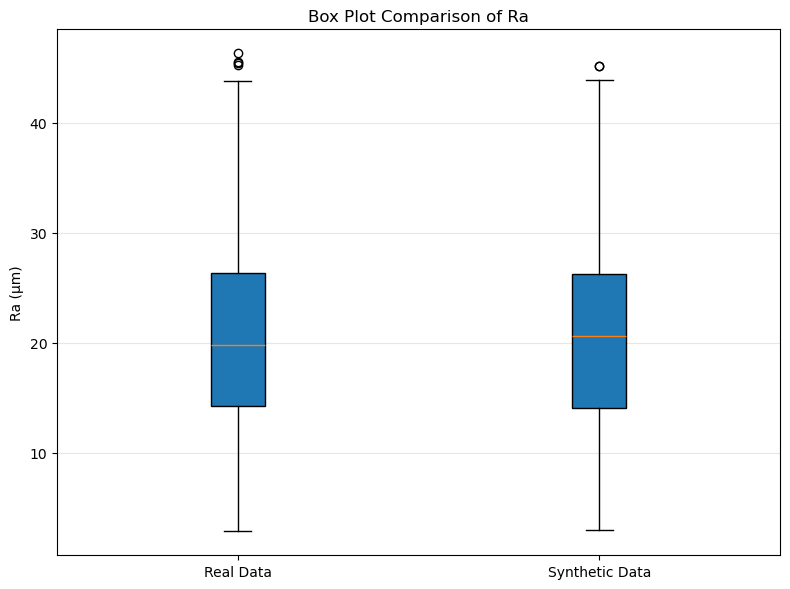}
    \caption{Box-plot comparison of $Ra$ values between real training data and CGAN-generated synthetic samples.}
    \label{FIG:15}
\end{figure}

To examine whether generated samples preserve the target distribution, Figure~\ref{FIG:14} and Figure~\ref{FIG:15} compare real and synthetic $Ra$ values used during training.

\subsection{SHAP results}
\label{subsec4.3}

To enhance the transparency of the selected MLP predictor, which was trained on both real and synthetic data, Shapley Additive Explanations (SHAP) were employed to quantify the contribution of each input feature to the predicted $Ra$ values. SHAP offers a consistent, model-agnostic attribution framework grounded in cooperative game theory, assigning each feature an additive value that reflects its influence on the prediction relative to a baseline expectation. All SHAP analyses in this study were conducted on the hold-out test set to facilitate interpretation of model behavior on previously unseen experimental data.

\begin{figure}[pos=H]
    \centering 
    \includegraphics[width=\linewidth]{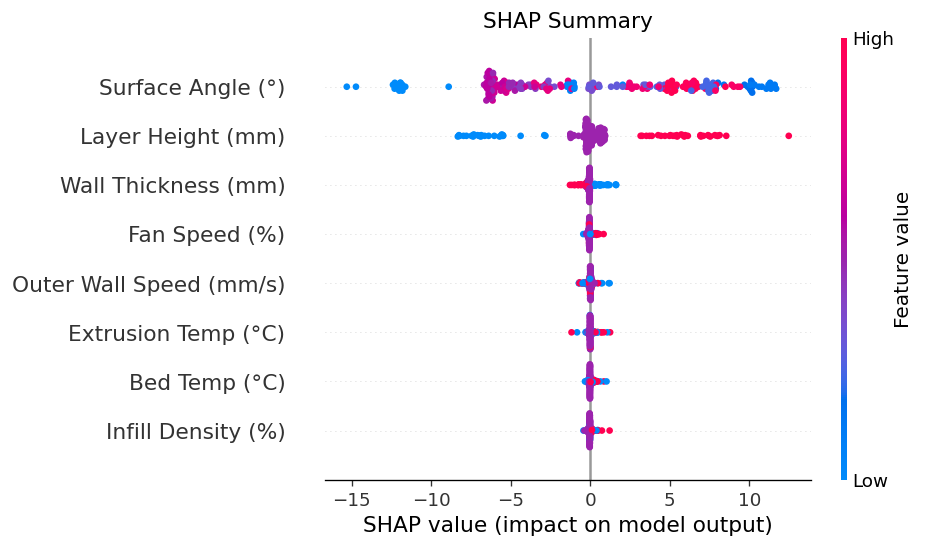}
    \caption{SHAP summary plot on the hold-out test set for the selected MLP model.}
    \label{FIG:16}
\end{figure}

\begin{figure}[pos=H]   
    \centering 
    \includegraphics[width=\linewidth]{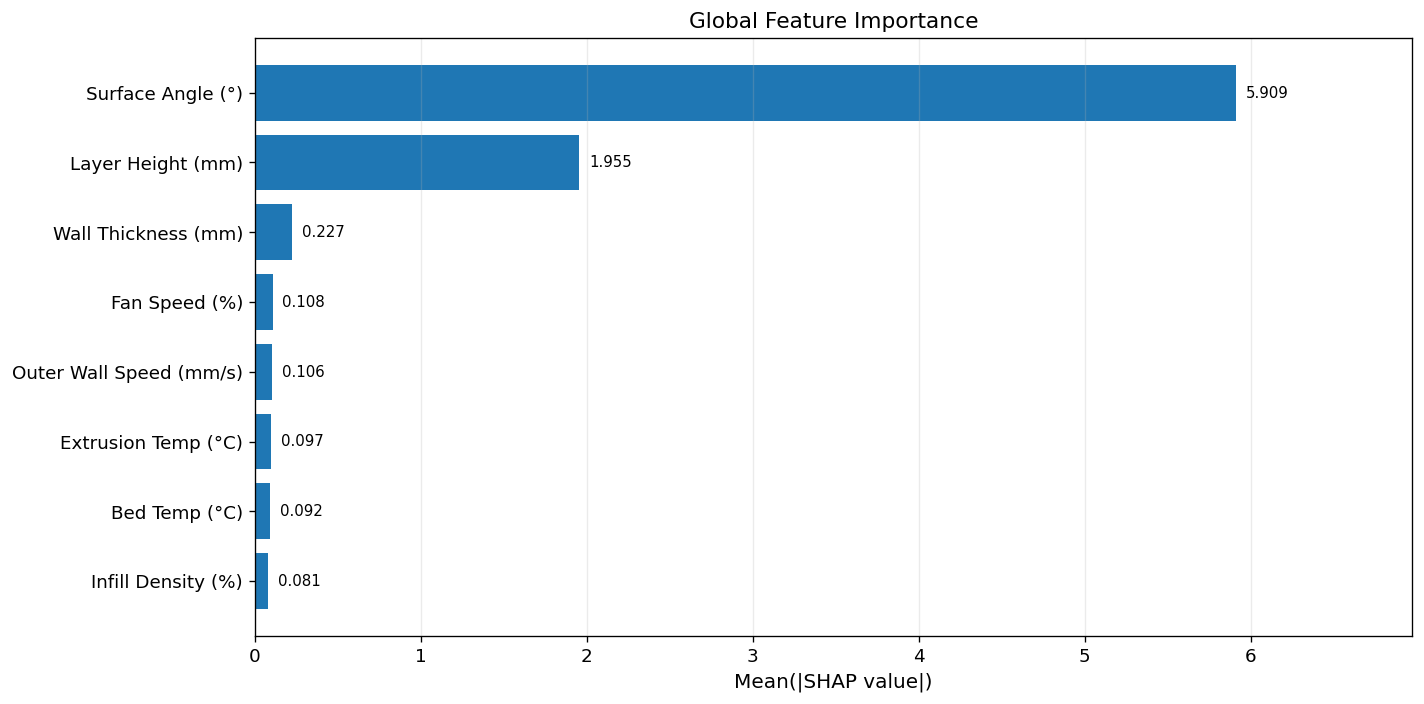} \caption{Global feature importance on the hold-out test set for the selected MLP model, computed as mean absolute SHAP values.} \label{FIG:17} 
\end{figure}

Figure~\ref{FIG:16} displays the SHAP summary (beeswarm) plot, which illustrates the distribution of feature contributions across all test samples. Each point represents a single test observation; the horizontal axis indicates whether the feature increased or decreased the predicted surface roughness, and the color denotes the magnitude of the feature value from low to high. Figure~\ref{FIG:17} presents the global feature importance ranking, obtained by aggregating SHAP values across the test set.

\section{Discussion}
\label{sec5}

This section interprets the predictive performance of an MLP regressor trained on experimental data and evaluates the effect of CGAN-based tabular augmentation on generalization to an unchanged hold-out test set. Model interpretability is further assessed using SHAP to determine whether the learned relationships correspond with established physical drivers of FFF surface roughness. The discussion then connects predictive performance to practical pre-print decision-making using the developed interactive visualization interface.

The MLP trained solely on experimentally measured samples achieved MAE = 2.20~$\mu$m, MSE = 10.737~$\mu$m$^2$, $R^2 = 0.844$, and MAPE = 14.19\% on the hold-out test set (Figure~\ref{FIG:12}). The measured versus predicted scatter is concentrated around the ideal $y=x$ line, which demonstrates that the model can capture nonlinear relationships among process parameters, surface angle, and $Ra$ within the scope of the available experimental data.

A systematic increase in dispersion is observed at higher $Ra$ values, indicating greater prediction uncertainty for rougher surfaces. This trend aligns with the characteristics of FFF processes, where high-roughness regimes often result from more pronounced coupled effects, such as increased stair-stepping severity at specific inclinations and resolution-related factors, such as layer height. Additionally, these regimes are typically underrepresented in the dataset, which increases variance in models that rely solely on data-driven approaches.

Augmenting the real training set with CGAN-generated tabular samples, while maintaining an unchanged hold-out test set, resulted in improved predictive accuracy: MAE = 1.752~$\mu$m, MSE = 6.900~$\mu$m$^2$, $R^2 = 0.900$, and MAPE = 10.89\% (Figure~\ref{FIG:13}). The resulting scatter plot is more compact around the ideal line, which indicates a reduction in error magnitude and enhanced consistency on previously unseen real measurements.

These results support the hypothesis that conditional augmentation mitigates limited coverage in the joint space defined by process parameters and surface inclination. Specifically, augmentation reduces estimator variance by enriching sparsely sampled regions and smoothing the learned mapping, which benefits neural regressors that are sensitive to sample density in tabular data. Since model selection and reporting are performed using an unchanged hold-out test split, the observed improvement reflects enhanced generalization rather than an evaluation artifact.

The plausibility of the synthetic data was assessed by comparing the marginal distribution of generated $Ra$ values with those of the real training targets. The density comparison in Figure~\ref{FIG:14} demonstrates a close alignment between real and synthetic distributions across the dominant modes. Additionally, the box plot in Figure~\ref{FIG:15} reveals similar central tendency and spread. These diagnostics indicate that the CGAN generated plausible target ranges and largely preserved the global variability of the measured roughness signal.

However, downstream predictive improvements require not only alignment of the marginal distribution of $Ra$, but also preservation of conditional relationships between inputs, such as process parameters and surface inclination, and the target variable. The observed improvement in hold-out performance provides empirical evidence that the generated samples maintained sufficient condition-consistency to enhance MLP training for the evaluated test distribution.

To evaluate the physical credibility and decision-support suitability of the learned relationships, SHAP analysis was conducted on the selected MLP predictor (Figures~\ref{FIG:16}--\ref{FIG:17}). The global ranking, determined by mean absolute SHAP values, identifies Surface Angle as the primary contributor, followed by Layer Height. This ordering is consistent with established FFF surface formation mechanisms: inclination determines the severity and manifestation of stair-stepping and deposition geometry, whereas layer height directly influences vertical discretization and the prominence of layer steps.

The SHAP summary plot (Figure~\ref{FIG:16}) further demonstrates that the influence of surface inclination is not strictly monotonic across the test samples. Both positive and negative SHAP values are observed for different angle levels, indicating a geometry-dependent response rather than a single linear trend. In contrast, layer height displays a more consistent effect, with higher values generally contributing to increased predicted roughness. This observation is consistent with the anticipated rise in staircase amplitude as layer thickness increases.

The SHAP results also offer a mechanistic explanation for the effectiveness of conditional augmentation in improving generalization. When surface inclination serves as the primary factor but is influenced by process settings, underrepresented combinations of inclination and parameter levels may increase uncertainty within an exclusively experimental training set. Synthetic data enrichment can address these gaps, allowing the MLP to learn smoother and more stable conditional responses that generalize more effectively to unseen measurements.

The achieved accuracy levels demonstrate that surface roughness can be estimated with high fidelity using a compact set of process parameters and a geometry-derived inclination descriptor. The observed reduction in MAE following augmentation indicates that predictive quality can be improved without additional physical measurements, provided that augmentation is validated through strict protocols and an unchanged hold-out test set.

In addition to numerical performance, the developed web-based visualization interface implements the predictor as a decision-support tool. By integrating user-defined process settings with mesh-derived surface inclinations, the system provides real-time previews of predicted roughness as a color-mapped overlay on the uploaded geometry. Interactive updates in response to parameter changes and orientation adjustments facilitate rapid exploration of trade-offs, thereby reducing reliance on costly trial-and-error cycles during process planning.

Several limitations must be considered when interpreting these results. The dataset was collected using a fixed machine and material configuration, which enhances internal consistency but may limit direct applicability to other printers, nozzle diameters, slicer implementations, or filament types without further calibration. Furthermore, although CGAN augmentation improved generalization in the evaluated setting, synthetic generators only approximate the underlying data distribution. Therefore, augmentation should be regarded as a model component requiring validation, rather than as an unlimited substitute for experimental measurement.

\section{Conclusion}
\label{sec6}

This study introduced a data-driven framework for predicting arithmetic mean surface roughness in fused filament fabrication prior to printing. The framework integrates a structured design of experiments, standardized surface metrology, supervised learning, conditional data augmentation, and interactive visualization. This approach enables surface-quality-aware process planning using a concise set of controllable parameters and a geometry-derived surface inclination descriptor.

A primary contribution is the experimentally generated dataset, produced through systematic variation of process settings and surface angles combined with consistent profilometry measurements. This dataset provides a reproducible foundation for benchmarking roughness prediction in FFF and can serve as a reference for future modeling and comparative studies.

The predictive results demonstrate that an MLP regressor can learn the nonlinear mapping from printing parameters and surface inclination to $Ra$, achieving robust generalization on an unseen hold-out test set. Model interpretation supports physical plausibility; both SHAP analysis and performance trends confirm that Surface Angle and Layer Height are the primary determinants of surface roughness within the studied ranges, consistent with established effects of stair-stepping and resolution in layered manufacturing.

To mitigate data scarcity associated with metrology-intensive experimentation, CGAN augmentation was evaluated as a method to enrich the training distribution in a condition-aware manner. The augmented training strategy enhanced test-set accuracy relative to the non-augmented MLP, suggesting that synthetic enrichment can reduce generalization error under rigorous validation and hold-out evaluation. Distribution diagnostics demonstrated that generated targets were broadly consistent with the real roughness distribution, supporting the plausibility of the synthetic samples used during training.

In addition to predictive accuracy, the study implemented the neural model as a practical decision-support tool via a web-based GUI. Designers are able to upload 3D models, specify printing parameters, and visualize predicted roughness as a colormap mapped onto the mesh. The interactive modification of build orientation and parameter settings facilitates rapid exploration of trade-offs and reduces dependence on costly trial-and-error during process planning.

\section{Future work}
\label{sec7}

Several extensions could enhance the scope and industrial relevance of the proposed framework. First, the current approach may be expanded beyond planar facets to encompass more complex geometries by incorporating curved surfaces and more detailed local descriptors, such as curvature or neighborhood-based orientation statistics. This would enable prediction for intricate, non-planar components.

Second, generalizability should be evaluated across a broader range of materials and filament formulations. Testing additional polymers, such as acrylonitrile butadiene styrene (ABS) and polyethylene terephthalate glycol (PETG), as well as reinforced filaments (e.g., carbon-fiber composites), would clarify material-dependent effects and support more robust deployment in diverse application contexts.

Third, the feature space could be expanded by incorporating additional system-specific variables, such as printer model, nozzle diameter, and selected slicer settings. These extensions would help quantify cross-platform variability and support transferring the predictor across different printing setups with minimal recalibration.

Fourth, the methodology could be adapted to other additive manufacturing processes, such as stereolithography or selective laser sintering, by redefining process-appropriate input variables and metrology protocols while maintaining consistent evaluation and deployment principles.

Finally, integrating in-situ sensing and real-time monitoring, such as thermal, vibration, or optical signals, could enable online roughness estimation and closed-loop control. This advancement would transition the framework from pre-print decision support to adaptive process control, thereby improving reliability under time-varying disturbances and machine-specific dynamics.

\section*{Funding}
This research did not receive any specific grant from funding agencies in the public, commercial, or not-for-profit sectors.

\section*{Declaration of competing interest}
The authors declare that they have no known competing financial interests or personal relationships that could have appeared to influence the work reported in this paper.

\section*{Data availability}
The dataset used in this study is publicly available on Zenodo: https://doi.org/10.5281/zenodo.18827518.

\section*{CRediT authorship contribution statement}
\textbf{Engin Deniz Erkan:} Conceptualization, Methodology, Software, Data curation, Investigation, Formal analysis, Validation, Writing -- original draft, Visualization.
\textbf{Elif Surer:} Conceptualization, Methodology, Supervision, Project administration, Resources, Writing -- review \& editing.
\textbf{Ulas Yaman:} Conceptualization, Methodology, Supervision, Project administration, Resources, Writing -- review \& editing.

\appendix
\section{Experimental parameter matrix}

For completeness and reproducibility, this appendix reports the full set of process-parameter combinations used to fabricate the experimental specimens under the Box--Behnken design. Table~\ref{tbl4} lists the printing parameters associated with each specimen identifier.

\onecolumn

\begin{small}
\setlength\LTleft{0pt}
\setlength\LTright{0pt}
\begin{longtable}{@{\extracolsep{\fill}}cccccccc@{}}
    \caption{Experimental 3D Printing Parameters} \label{tbl4} \\
    
    \toprule
    \textbf{ID} & \textbf{Layer} & \textbf{Extrusion} & \textbf{Outer Wall} & \textbf{Infill} & \textbf{Wall} & \textbf{Bed} & \textbf{Fan} \\
     & \textbf{Height} & \textbf{Temp.} & \textbf{Speed} & \textbf{Density} & \textbf{Thickness} & \textbf{Temp.} & \textbf{Speed} \\
     & (mm) & ($^\circ$C) & (mm/s) & (\%) & (mm) & ($^\circ$C) & (\%) \\
    \midrule
    \endfirsthead

    \multicolumn{8}{c}{{\tablename\ \thetable{} -- continued from previous page}} \\
    \toprule
    \textbf{ID} & \textbf{Layer} & \textbf{Extrusion} & \textbf{Outer Wall} & \textbf{Infill} & \textbf{Wall} & \textbf{Bed} & \textbf{Fan} \\
     & \textbf{Height} & \textbf{Temp.} & \textbf{Speed} & \textbf{Density} & \textbf{Thickness} & \textbf{Temp.} & \textbf{Speed} \\
     & (mm) & ($^\circ$C) & (mm/s) & (\%) & (mm) & ($^\circ$C) & (\%) \\
    \midrule
    \endhead

    \bottomrule
    \multicolumn{8}{r}{{Continued on next page...}} \\
    \endfoot

    \bottomrule
    \endlastfoot

    Object-1 & 0.12 & 190 & 200 & 15 & 0.42 & 60 & 80 \\
    Object-2 & 0.12 & 210 & 200 & 15 & 0.42 & 60 & 80 \\
    Object-3 & 0.28 & 190 & 200 & 15 & 0.42 & 60 & 80 \\
    Object-4 & 0.28 & 210 & 200 & 15 & 0.42 & 60 & 80 \\
    Object-5 & 0.12 & 200 & 150 & 15 & 0.42 & 60 & 80 \\
    Object-6 & 0.12 & 200 & 250 & 15 & 0.42 & 60 & 80 \\
    Object-7 & 0.28 & 200 & 150 & 15 & 0.42 & 60 & 80 \\
    Object-8 & 0.28 & 200 & 250 & 15 & 0.42 & 60 & 80 \\
    Object-9 & 0.12 & 200 & 200 & 5 & 0.42 & 60 & 80 \\
    Object-10 & 0.12 & 200 & 200 & 25 & 0.42 & 60 & 80 \\
    Object-11 & 0.28 & 200 & 200 & 5 & 0.42 & 60 & 80 \\
    Object-12 & 0.28 & 200 & 200 & 25 & 0.42 & 60 & 80 \\
    Object-13 & 0.12 & 200 & 200 & 15 & 0.36 & 60 & 80 \\
    Object-14 & 0.12 & 200 & 200 & 15 & 0.48 & 60 & 80 \\
    Object-15 & 0.28 & 200 & 200 & 15 & 0.36 & 60 & 80 \\
    Object-16 & 0.28 & 200 & 200 & 15 & 0.48 & 60 & 80 \\
    Object-17 & 0.12 & 200 & 200 & 15 & 0.42 & 55 & 80 \\
    Object-18 & 0.12 & 200 & 200 & 15 & 0.42 & 65 & 80 \\
    Object-19 & 0.28 & 200 & 200 & 15 & 0.42 & 55 & 80 \\
    Object-20 & 0.28 & 200 & 200 & 15 & 0.42 & 65 & 80 \\
    Object-21 & 0.12 & 200 & 200 & 15 & 0.42 & 60 & 60 \\
    Object-22 & 0.12 & 200 & 200 & 15 & 0.42 & 60 & 100 \\
    Object-23 & 0.28 & 200 & 200 & 15 & 0.42 & 60 & 60 \\
    Object-24 & 0.28 & 200 & 200 & 15 & 0.42 & 60 & 100 \\
    Object-25 & 0.2 & 190 & 150 & 15 & 0.42 & 60 & 80 \\
    Object-26 & 0.2 & 190 & 250 & 15 & 0.42 & 60 & 80 \\
    Object-27 & 0.2 & 210 & 150 & 15 & 0.42 & 60 & 80 \\
    Object-28 & 0.2 & 210 & 250 & 15 & 0.42 & 60 & 80 \\
    Object-29 & 0.2 & 190 & 200 & 5 & 0.42 & 60 & 80 \\
    Object-30 & 0.2 & 190 & 200 & 25 & 0.42 & 60 & 80 \\
    Object-31 & 0.2 & 210 & 200 & 5 & 0.42 & 60 & 80 \\
    Object-32 & 0.2 & 210 & 200 & 25 & 0.42 & 60 & 80 \\
    Object-33 & 0.2 & 190 & 200 & 15 & 0.36 & 60 & 80 \\
    Object-34 & 0.2 & 190 & 200 & 15 & 0.48 & 60 & 80 \\
    Object-35 & 0.2 & 210 & 200 & 15 & 0.36 & 60 & 80 \\
    Object-36 & 0.2 & 210 & 200 & 15 & 0.48 & 60 & 80 \\
    Object-37 & 0.2 & 190 & 200 & 15 & 0.42 & 55 & 80 \\
    Object-38 & 0.2 & 190 & 200 & 15 & 0.42 & 65 & 80 \\
    Object-39 & 0.2 & 210 & 200 & 15 & 0.42 & 55 & 80 \\
    Object-40 & 0.2 & 210 & 200 & 15 & 0.42 & 65 & 80 \\
    Object-41 & 0.2 & 190 & 200 & 15 & 0.42 & 60 & 60 \\
    Object-42 & 0.2 & 190 & 200 & 15 & 0.42 & 60 & 100 \\
    Object-43 & 0.2 & 210 & 200 & 15 & 0.42 & 60 & 60 \\
    Object-44 & 0.2 & 210 & 200 & 15 & 0.42 & 60 & 100 \\
    Object-45 & 0.2 & 200 & 150 & 5 & 0.42 & 60 & 80 \\
    Object-46 & 0.2 & 200 & 150 & 25 & 0.42 & 60 & 80 \\
    Object-47 & 0.2 & 200 & 250 & 5 & 0.42 & 60 & 80 \\
    Object-48 & 0.2 & 200 & 250 & 25 & 0.42 & 60 & 80 \\
    Object-49 & 0.2 & 200 & 150 & 15 & 0.36 & 60 & 80 \\
    Object-50 & 0.2 & 200 & 150 & 15 & 0.48 & 60 & 80 \\
    Object-51 & 0.2 & 200 & 250 & 15 & 0.36 & 60 & 80 \\
    Object-52 & 0.2 & 200 & 250 & 15 & 0.48 & 60 & 80 \\
    Object-53 & 0.2 & 200 & 150 & 15 & 0.42 & 55 & 80 \\
    Object-54 & 0.2 & 200 & 150 & 15 & 0.42 & 65 & 80 \\
    Object-55 & 0.2 & 200 & 250 & 15 & 0.42 & 55 & 80 \\
    Object-56 & 0.2 & 200 & 250 & 15 & 0.42 & 65 & 80 \\
    Object-57 & 0.2 & 200 & 150 & 15 & 0.42 & 60 & 60 \\
    Object-58 & 0.2 & 200 & 150 & 15 & 0.42 & 60 & 100 \\
    Object-59 & 0.2 & 200 & 250 & 15 & 0.42 & 60 & 60 \\
    Object-60 & 0.2 & 200 & 250 & 15 & 0.42 & 60 & 100 \\
    Object-61 & 0.2 & 200 & 200 & 5 & 0.36 & 60 & 80 \\
    Object-62 & 0.2 & 200 & 200 & 5 & 0.48 & 60 & 80 \\
    Object-63 & 0.2 & 200 & 200 & 25 & 0.36 & 60 & 80 \\
    Object-64 & 0.2 & 200 & 200 & 25 & 0.48 & 60 & 80 \\
    Object-65 & 0.2 & 200 & 200 & 5 & 0.42 & 55 & 80 \\
    Object-66 & 0.2 & 200 & 200 & 5 & 0.42 & 65 & 80 \\
    Object-67 & 0.2 & 200 & 200 & 25 & 0.42 & 55 & 80 \\
    Object-68 & 0.2 & 200 & 200 & 25 & 0.42 & 65 & 80 \\
    Object-69 & 0.2 & 200 & 200 & 5 & 0.42 & 60 & 60 \\
    Object-70 & 0.2 & 200 & 200 & 5 & 0.42 & 60 & 100 \\
    Object-71 & 0.2 & 200 & 200 & 25 & 0.42 & 60 & 60 \\
    Object-72 & 0.2 & 200 & 200 & 25 & 0.42 & 60 & 100 \\
    Object-73 & 0.2 & 200 & 200 & 15 & 0.36 & 55 & 80 \\
    Object-74 & 0.2 & 200 & 200 & 15 & 0.36 & 65 & 80 \\
    Object-75 & 0.2 & 200 & 200 & 15 & 0.48 & 55 & 80 \\
    Object-76 & 0.2 & 200 & 200 & 15 & 0.48 & 65 & 80 \\
    Object-77 & 0.2 & 200 & 200 & 15 & 0.36 & 60 & 60 \\
    Object-78 & 0.2 & 200 & 200 & 15 & 0.36 & 60 & 100 \\
    Object-79 & 0.2 & 200 & 200 & 15 & 0.48 & 60 & 60 \\
    Object-80 & 0.2 & 200 & 200 & 15 & 0.48 & 60 & 100 \\
    Object-81 & 0.2 & 200 & 200 & 15 & 0.42 & 55 & 60 \\
    Object-82 & 0.2 & 200 & 200 & 15 & 0.42 & 55 & 100 \\
    Object-83 & 0.2 & 200 & 200 & 15 & 0.42 & 65 & 60 \\
    Object-84 & 0.2 & 200 & 200 & 15 & 0.42 & 65 & 100 \\
    Object-85 & 0.2 & 200 & 200 & 15 & 0.42 & 60 & 80 \\
    Object-86 & 0.2 & 200 & 200 & 15 & 0.42 & 60 & 80 \\
    Object-87 & 0.2 & 200 & 200 & 15 & 0.42 & 60 & 80 \\
    \end{longtable}
\end{small}

\twocolumn

\printcredits

\bibliographystyle{elsarticle-num}

\bibliography{cas-refs}

\end{document}